\documentclass[letterpaper, 10 pt, conference]{IEEEconf}  

\usepackage{lmodern}
\usepackage[T1]{fontenc}

\IEEEoverridecommandlockouts                              
\usepackage{graphics} 
\usepackage{epsfig} 
\usepackage{mathptmx} 
\usepackage{times}
\usepackage{amsmath} 
\usepackage{amssymb} 
\usepackage{xcolor} 
\usepackage{subcaption}
\usepackage{wrapfig}
\usepackage{algorithm}
\usepackage{algpseudocode}
\usepackage{url}
\usepackage{caption}
\usepackage{subcaption}
\usepackage{epstopdf}
\usepackage{times}

\title{\LARGE \bf
Experimental Setup and Software Pipeline to Evaluate Optimization based Autonomous Multi-Robot Search Algorithms
}

\author{Aditya Bhatt$^{1}$, Mary Katherine Corra$^{2}$, Franklin Merlo$^{1}$, Prajit KrisshnaKumar$^{1}$ and Souma Chowdhury$^\dagger$ 
\thanks{$^\dagger$ Corresponding Author, soumacho@buffalo.edu}
\thanks{$^{1}$ Department of Mechanical And Aerospace Engineering,
        University at Buffalo, Buffalo, NY 
        }%
\thanks{$^{2}$ Department of Computer Science and Engineering,
University at Buffalo, Buffalo, NY }%
\thanks{*This work was supported by the NSF award CMMI 2048020, and undergraduate funding from IAD at University at Buffalo. Any opinions, findings, conclusions, or recommendations expressed in this paper are those of the authors and do not necessarily reflect the views of NSF or IAD.}
\thanks{Copyright \textcopyright 2025 ASME. Personal use of this material is permitted. Permission from ASME must be obtained for all other uses, in any current or future media, including reprinting/republishing this material for advertising or promotional purposes, creating new collective works, for resale or redistribution to servers or lists, or reuse of any copyrighted component of this work in other works}
}

\begin{document}

\maketitle
\thispagestyle{empty}
\pagestyle{empty}

\begin{abstract}

Signal source localization has been a problem of interest in the multi-robot systems domain given its applications in search \& rescue and hazard localization in various industrial and outdoor settings. A variety of multi-robot search algorithms exist that usually formulate and solve the associated autonomous motion planning problem as a heuristic model-free or belief model-based optimization process. Most of these algorithms however remains tested only in simulation, thereby losing the opportunity to generate knowledge about how such algorithms would compare/contrast in a real physical setting in terms of search performance and real-time computing performance, and their potential to later translate to field deployment. To address this gap, this paper presents a new lab-scale physical setup and associated open-source software pipeline to evaluate and benchmark multi-robot search algorithms. The presented physical setup innovatively uses an acoustic source (that is safe and inexpensive) and small ground robots (e-pucks) operating in a standard motion-capture environment. This setup can be easily recreated and used by most robotics researchers. The acoustic source also presents interesting uncertainty in terms of its noise-to-signal ratio, which is useful to assess sim-to-real gaps. The overall software pipeline is designed to readily interface with any multi-robot search algorithm with minimal effort and is executable in parallel asynchronous form. This pipeline includes a framework for distributed implementation of multi-robot or swarm search algorithms, integrated with a ROS (Robotics Operating System)-based software stack for motion capture supported localization. Source calibration, signal filtering and robot communication solutions that are key to effective use of such a setup are also presented. The utility of this novel setup is demonstrated by using it to evaluate two state-of-the-art multi-robot search algorithms, based on swarm optimization and batch-Bayesian Optimization (called Bayes-Swarm), as well as a random walk baseline. A comparative analysis of their performance is then provided, highlighting the performance superiority of Bayes-Swarm. Finally, practical algorithmic improvements are also facilitated by the experiments.

\end{abstract}

\section{Introduction}
The problem of signal source localization is recurring in robotics and can be found in applications such as disaster response \cite{tadokoro2019disaster}, gas leakage source localization \cite{baetz2009mobile}, radio source localization \cite{charrow2014cooperative, viseras2016decentralized}, target search \cite{behjat2021learning} and acoustic source localization for search and rescue missions \cite{8490881}. 
In such time-sensitive critical missions, using a team of robots which can collaborate has several advantages over a single robot \cite{lochmatter2009understanding, tan2013research}. 
The research efforts on multi-robot search algorithms aiming to exploit these advantages have been proposed across domains of optimization \cite{wang2022odor, DADGAR201662}, control \cite{sinha2018convergence} or probabilistic inference \cite{jing2021recent, ghassemi2020extended}.
Particularly, model-based batch optimization algorithms (e.g., Bayesian Optimization) and population-based optimization algorithms (e.g., particle swarm concepts) have been recently translated from usual non-embodied design optimization applications to provide decision-support in operation of autonomous multi-robotic or swarm systems (MRSS) \cite{ghassemi2022penalized, DADGAR201662}.

Nevertheless, just like in the case of design optimization, there is often a significant sim-to-real gap in performance -- i.e., performance observed in simulation could vary significantly from performance in physical world. This is usually because of the difference in signal distribution observed in real world versus the signal distribution used in simulation. The signal distribution in simulation is modeled using an analytical function (usually Gaussian\cite{ghassemi2022penalized}), while the real world signal distribution has a lot of interference. Further, motion and communication uncertainties also add to the sim-to-real gap.  
\textit{Hence, there's a growing need for testbeds and experimental protocols to understand, quantify and address such sim-to-real gaps in optimization driven MRSS autonomy. 
As a first step towards addressing this need, we present a safe and easy-to-recreate physical setup and associated software pipeline to evaluate swarm search algorithms in a lab-scale physical experimental environment. Our goal is to make the hardware setup blueprint and software stack available to the broader robotics community allowing other researchers to readily create and use them for evaluating and benchmarking their multi-robot/swarm search algorithms.} While the formal systematic analysis of sim-to-real gaps, their quantification and mitigation addressal is not within the scope of this paper, the methods and pipelines presented here is an important first step towards achieving those goals. 

In the remainder of this introductory section, we provide the related work for development of such testbeds, followed by the main contributions and a summary of the remaining sections.

A survey on small-scale testbeds for connected and automated vehicles and robot swarms can be found in \cite{mokhtarian2024survey}. An open source UAV platform for swarm robotics was proposed in \cite{ouguz2024open}. With regards to Multi-Robot signal source localization, individual research groups have used their own setups for signal source localization. For example, two robots were used to localize a third robot in \cite{charrow2014cooperative} using radio based time of flight range sensors. For the evaluation of their gas source localization algorithm, Wiedemann et al \cite{wiedemann2019model} conducted lab-scale experiments with ethanol gas and a swarm of robots. The existing setups for multi-robot search algorithms lack one of the following three desirable characteristics: \textit{a)} an inexpensive setup, \textit{b)} a setup that is built on systems/capabilities that are usually available in any mainstream robotics lab or robotics program and \textit{c)} a setup that is safe to deploy. 
In this paper, we have developed an experimental indoor setup with acoustic signal source. This setup uses a motion-capture system to get real-time global co-ordinates of the robot. An acoustic signal source is used which generates sound signals with certain frequency at some intensity, and the goal of the multi-robot team is to localize the source as quickly as possible based on the sound measurements collected as they move around in that 2D environment.  
Figure \ref{fig:cad-expt} shows our lab experiment setup to implement search algorithms on small ground robots. The positions of the robots are obtained using the motion capture cameras (shown on the wall). Our test environment (table in the center) has dimensions of $2 \text{m} \times 2 \text{m}$.

The main contributions of this paper can thus be summarized as: 
\textbf{1)} Development of a relatively inexpensive and safe physical setup to run multi-robot signal source localization experiments using an acoustic source and small two-wheeled ground-robots with onboard microphones. 
\textbf{2)} Development of a Robotic Operating System (ROS) \cite{quigley2009ros} based  software stack for implementing swarm search algorithms on physical robots in a parallel asynchronous manner, supported by noise filtering and navigation algorithms.   
\textbf{3}) Implementation and demonstration of the hardware/software setup with three different multi-robot search algorithms, namely Bayes-Swarm, Swarm Optimization and random walk; and improvement of Bayes-Swarm based on observed real experimental performance.

In the next section, we have given the problem definition and execution of a generic multi-robot search algorithm. Following that, in section \ref{sec:exp-setup}, we have explained in detail the experimental setup and our ROS based software pipeline. Section \ref{sec:swarm-search-algos} then gives a brief overview of existing swarm search algorithms, along with mathematical details of the algorithms we deployed in our setup. Subsequently, we present the experiments and our findings in section \ref{sec:num-exps}. The paper ends with our concluding remarks and the future work, given in section \ref{sec:concl}.

\section{Multi-Robot Search Problem Definition} \label{sec:search-pb-def}
Here onwards, we will use the term \textit{swarm search} to refer to the process of multi-robot search to localize the signal source of maximum strength in a given environment (over which the signal varies). In this paper we specifically consider a 2D acoustic signal environment with a single sound source.

Assume that the signal distribution is governed by the function $\phi(x, y)$, where $(x, y)$ represents the location coordinates. The problem of signal source localization can then be mathematically expressed as an optimization problem with the following objective equation:
\begin{equation} \label{eqn:obj-func}
    \mathbf{(\bar{x}, \bar{y})} =  {\arg \max} \hspace{1mm} \phi(x, y), ~{\text{s.t.}}~ [x_{\text{L}},y_{\text{L}}] \le [x,y] \le [x_{\text{U}},y_{\text{U}}]
\end{equation}

Here, $[x_{\text{L}},y_{\text{L}}]$ and $[x_{\text{U}},y_{\text{U}}]$ are the bounds of the search arena, $\phi(x, y) \gets$ \textit{Signal Distribution Function} and $\mathbf{(\bar{x}, \bar{y})}$ represents the source location. The goodness of a multi-robot search algorithm depends on both: 1) how closely the above optimal search can be solved (within a prescribed time for an application, at least one of the robots should be able to converge to the location $\mathbf{(\bar{x}, \bar{y})}$) and 2) how long it takes in physical time to solve it. 


However, in practical search missions, an analytical form of $\phi$ is not available. Robots equipped with appropriate onboard sensors can sample or measure the signal function value $\phi(x_t, y_t)$ at any location $(x_t, y_t)$ at time $t$. They do so progressively in their quest to find or get close to the source location $\mathbf{(\bar{x}, \bar{y})}$. In this case, the physical time spent in such a search process is mainly attributed to robot motion between waypoints and (to a usually smaller extent) to the computing time spent in deciding the next waypoint (while idling at the current one). With multiple robots that share information with each other, the environment can be sampled more efficiently, where each robot can decide the next waypoint to visit or path to take based on the signal observations collected so far by itself and its peers (communicated to it). 

The process of solving the swarm search problem can thus be perceived as an informative path-planning method. Starting from a depot or spawning area, the initial path can be decided by each robot randomly or based on some heuristic. After the first waypoint, the robots need to make an informed decision for the next waypoint based on the data sampled by the robot along its trajectory, appended with data shared by its peers. A pseudocode of a generic execution of a swarm mission is shown in Algorithm \ref{alg:swarm-mission}. The termination criterion is usually a physical time limit. The mission is considered to be failed if the robots do not locate or get close enough to the signal source (target) within this given physical time limit.

\begin{algorithm}
\caption{Generic Swarm Search-Mission} \label{alg:swarm-mission}
\footnotesize
\begin{algorithmic}[1]
\State \textbf{Input:} Termination Criteria 
 \State $v_r \gets \text{robot speed}$ 
 \State $fs \gets \text{sensor sampling rate}$ 
\State Initiate mission with initial robot positions 
\State Generate Initial way-points $(x_r^{p}, y_r^{p})$
\State \textbf{robot-r} \algorithmiccomment{in parallel w.r.t all robots r$:1 \dots n_r$}
\While{Termination Criteria}
\State $(x_k, y_k, \phi_k)_{k=1}^{N_r} = \texttt{NavigationAlgorithm}(x_r^{p}, y_r^{p})$  
\State $\mathbf{D} = (x_i, y_i, \phi_i)_{i=1}^{N}$ \algorithmiccomment{$N = N_r (\text{sampled data}) + N_{-r} (\text{peers' data})$}
\State $(x_r^{p}, y_r^{p}) \gets \text{Swarm Search Algorithm}(\mathbf{D})$
\EndWhile

\Function{Navigation Algorithm}{$x_f, y_f$} 
\State $(x_f, y_f) \gets \text{Desired robot position}$
\State $(x_k, y_k) \gets \text{Motion-Capture}$ \algorithmiccomment{initial robot-position}
\State $d = ||(x_f, y_f) - (x_k, y_k)||$
\State traj-obs $\gets$ $\text{EmptyArray()}$  \algorithmiccomment{Data sampled along the trajectory}
\While{$||(x_f, y_f) - (x_k, y_k)|| \leq$ tol} 
\State $(x_k, y_k) \gets \text{Motion-Capture}$ \algorithmiccomment{current robot-position}
\State $\phi_k (x_k, y_k) \gets \text{sensor}$ \algorithmiccomment{signal-value at $(x_k, y_k)$} 
\State traj-obs.\texttt{append}($x_k, y_k, \phi_k$)
\State $(v_x, \omega) \gets \texttt{PID}((x_f, y_f), (x_k, y_k))$   \algorithmiccomment{linear($v_x$) \& angular ($\omega$) velocity}
\EndWhile 
\State $N_r \gets$ \texttt{length}(traj-obs) 
$= \frac{d}{v_r}.fs$ 
\State \textbf{return} $(x_k, y_k, \phi_k)_{k=1}^{N_r}$
\EndFunction
\end{algorithmic}
\end{algorithm}

\section{Experimental Setup} \label{sec:exp-setup}
\begin{figure}[h]
\centering
\begin{minipage}{\linewidth}
    \includegraphics[width=0.95\linewidth]{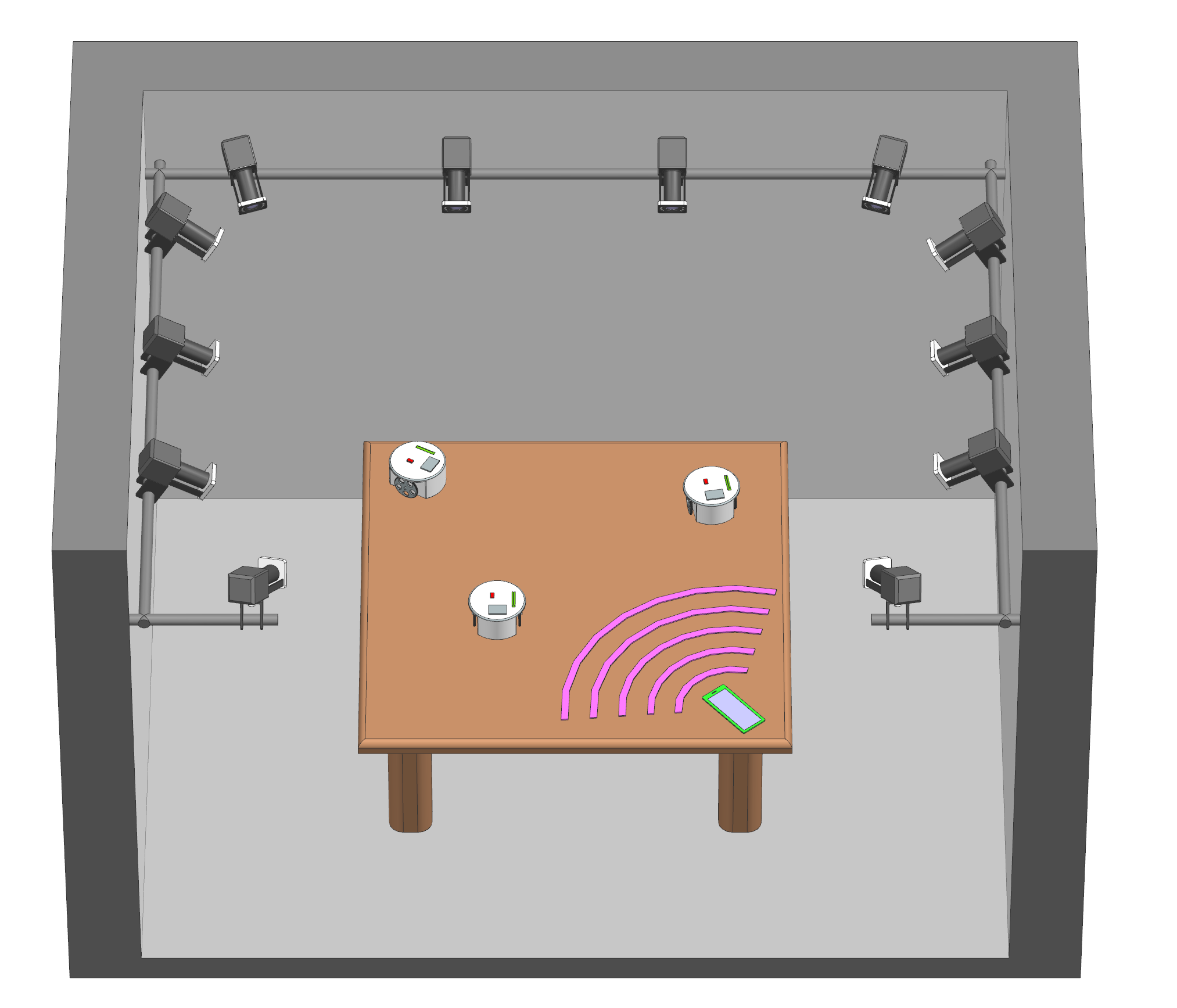}
    \subcaption{CAD model of the lab setup}\label{subfig:cad-setup}
\end{minipage}
    \begin{minipage}{\linewidth}
    \includegraphics[width=0.95\linewidth]{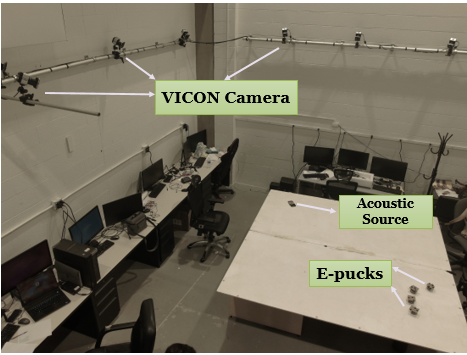}
    \subcaption{The lab setup} \label{subfig:lab-setup}
    \end{minipage}
    \caption{Indoor table-top testbed (2m $\times$ 2m) within motion capture space (cameras shown on wall) for experimental evaluation of multi-robot search algorithms.}
    \label{fig:cad-expt}
    \end{figure}

This section describes our new experimental setup to test swarm search algorithms.
This setup includes the following hardware components:
\textbf{1)} physical robots - e-pucks \textbf{2)} motion capture system and \textbf{3)} a central computing node. 
Note that the computational power of the e-puck robots (or similar small form factor swarm-bots) is not amenable for handling complex computations, especially those needed by model-based swarm search algorithms; therefore, in our setup, the major computations take place on the central computing node or workstation. 
Despite centralized computations, our software stack is designed to allow a decentralized implementation, where each robot makes its own decision and then communicates the required information to its peers. The software pipeline facilitates a \textit{plug and play setup} for any multi-robotic search algorithm to be evaluated. 
\textit{This software stack is made open-source as a github repository} \cite{adams_signal_search_testbed}.

Below we first describe the hardware components of our setup, followed by a description of the software and communication pipeline that we have developed and implemented to allow decentralized execution of swarm search algorithms. Lastly, in this section, we explain how the signal source is calibrated and how the signal observations recorded by robots are processed to be used in a search algorithm.

\subsection{Hardware} \label{sec:hardware}
\textbf{Robots:}
We use e-puck2 robots here, which are cylindrical, small robots with a 7 cm diameter. They have been used as a common platform in swarm robotics research \cite{ alkilabi2017cooperative, harmanda2021development, florea2020overview}. 
All the experiments in this study are conducted with 4 e-pucks. 
These e-pucks have various onboard sensors. Those employed in our experiments include four microphone (mic) sensors
placed on the front face, used to obtain acoustic signal measurements.
These robots operate on the 2 m $\times$ 2 m table-top testbed shown in Fig. \ref{fig:cad-expt}, with the sound source placed near one corner of the testbed. 

\textbf{Motion Capture System:}
A VICON motion capture system \cite{viconMotionCapture} comprising 12 cameras, a facility in our lab space within University at Buffalo, is used to localize the e-pucks. The testbed table is located $\sim$centrally relative to the motion capture cameras, to ensure appropriate motion capture coverage of the entire testbed surface. 
Each e-puck is equipped with three or four reflective markers (shown in Fig.~\ref{fig:exp-setup}) with different configurations and is created as an object in Vicon tracker software. This enables live tracking of the e-pucks with the locations of each e-puck being updated at 100 hz.

\textbf{Central Device - Laptop:} 
As stated previously, the onboard processor on the e-puck 2 robots (\texttt{esp-32} chip) is not capable of performing expensive computations. Therefore, all the computations were performed on a mobile workstation: Dell Inspiron 16, CPU: AMD Ryzen 7 7730u with 8GB RAM.

\textbf{Signal Source:} 
For simplicity and ease of adoption, our setup uses a cell phone -- POCO F1, Xiaomi -- as the acoustic signal source.  Using a frequency generator application, the sound signals were generated at a frequency of $573$ Hz. Higher fidelity sound sources, such as calibrated speakers, can also be used to produce robust signal environments. 

\textbf{Hardware Considerations:}
1) Extremely low ground clearance of the e-puck robot wheels can cause them to get stuck easily, and hence a relatively flat surface with low roughness is required for the experiments to progress well. 
2) The microphone sensors (similar to the ones on epucks) are commercially available and can be integrated with a custom robot as well. However, it is crucial to ensure that the noise generated by robot motors does not exceed the decibel range of sensors. 

\subsection{Communication} \label{sec:com}
\begin{figure}
    \centering
    \includegraphics[width=0.8\linewidth]{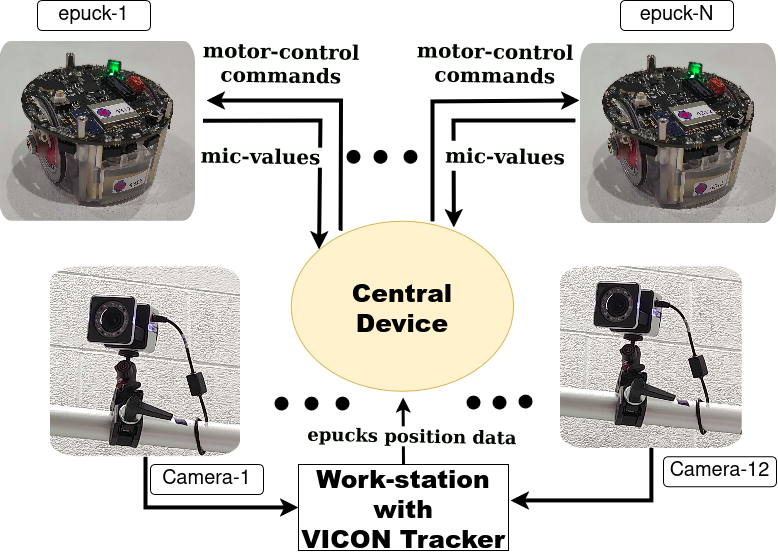}
    \caption{Experimental setup involving a central computing device, individual e-puck robots and motion capture system (incl. workstation), and how they exchange information}
    \label{fig:exp-setup}
\end{figure}

Figure \ref{fig:exp-setup} illustrates how the different hardware components: central device, robots and motion capture communicate with each other when executing a generic multi-robot search algorithm (pseudocode given in Alg.~\ref{alg:swarm-mission}). 
The inter-device communication occurs through TCP, where the central device receives mic-values from the e-pucks, position data from the motion capture system, and sends motor-control commands back to the e-puck robots. 


\begin{figure}
    \centering
    \includegraphics[width=\linewidth]{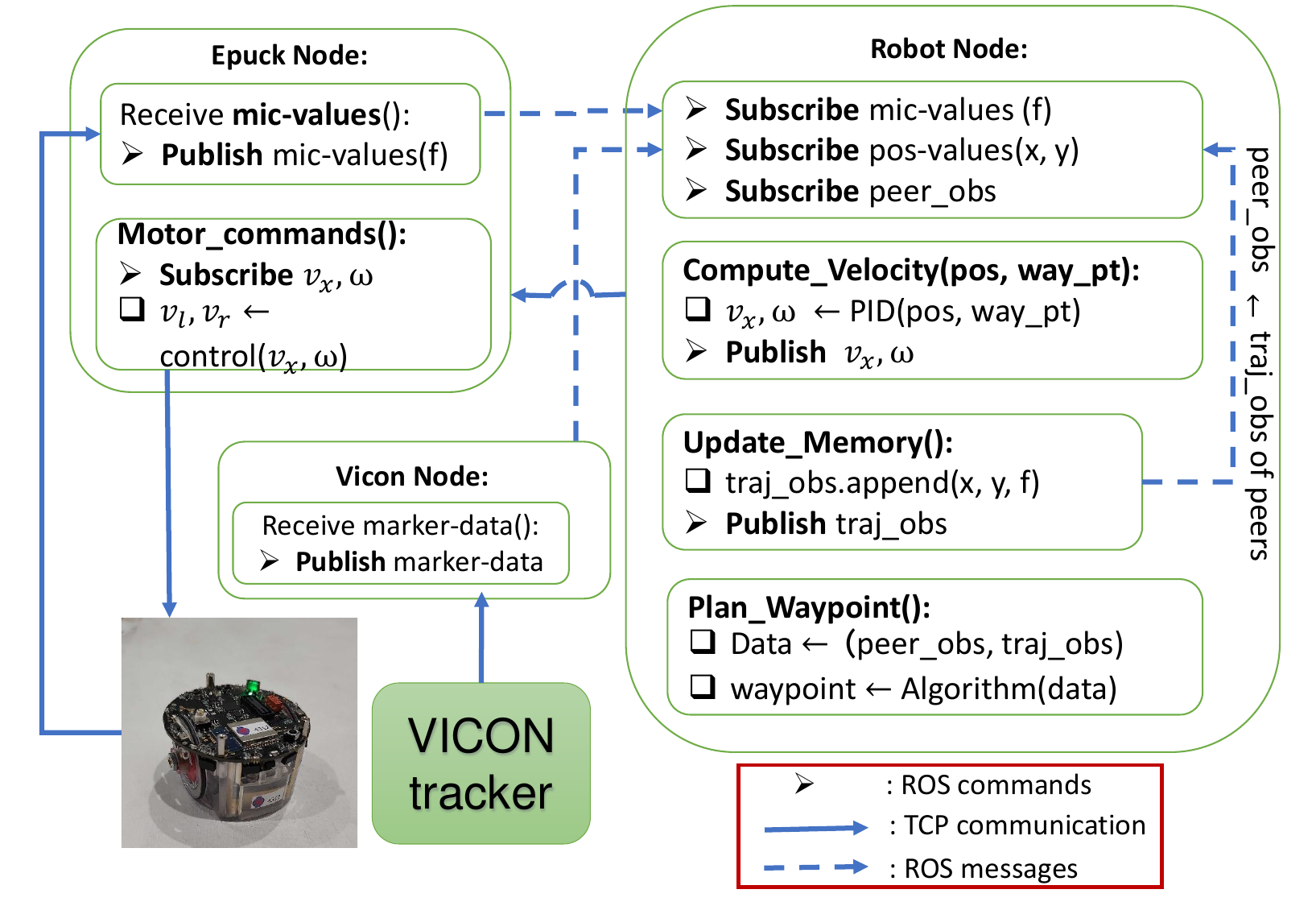}
    \caption{ROS based communication framework in our experimental setup. All the three nodes (epuck node, robot node and vicon node) are part of our software stack and are run on the workstation (dell laptop in our case). Epuck Node is responsible for communicating with epuck (hardware), Robot Node runs swarm-search algorithm and computes velocities, Vicon Node is responsible for obtaining the real-time robot location from motion-capture system}
    \label{fig:ros-com}
\end{figure}
Within the central device, we use ROS framework to communicate the mic-values and the position data with navigation and search algorithms.
For that purpose, we define three kind of ROS nodes: robot-hardware-node, robot-software-node, and vicon-node. Since we use e-pucks as the physical robots, robot-hardware-nodes are termed as epuck-nodes from now on and robot-software-nodes termed as robot-nodes from now on. 
The ROS package given by \cite{epuck-driver-cpp} creates a ROS-node for each e-puck (epuck-node) to receive mic-values from e-puck and send motor-velocity commands (through TCP - solid lines in Fig.~\ref{fig:ros-com}). 
The ROS package given by \cite{vicon} creates a ROS-node (vicon-node) which receives position-data (through TCP - solid lines in  Fig.~\ref{fig:ros-com}).  
Our software stack initializes a node for each robot (robot-node) which handles the execution of navigation and swarm search algorithms. 

To elaborate, the position-data is published by the vicon-node to a ROS-topic and the robot-nodes subscribe to this ROS-topic to receive the position data ($x_t, y_t$) (shown as dashed lines in  Fig.~\ref{fig:ros-com}). 
The mic-values are published by epuck-nodes to another ROS-topic which is then subscribed by the robot-nodes to receive the signal data $\phi(x_t, y_t)$ (shown as dashed lines in  Fig.~\ref{fig:ros-com}). 
The swarm search algorithms leverage parallelism in data collection and information sharing. Therefore, each robot has to broadcast the observations data that it recorded along its trajectory and any other information required by the algorithm. This information sharing among robots is enabled through ROS.
As shown in Fig.~\ref{fig:ros-com}, the robot-nodes publish the data to a ROS-topic which is then subscribed by its peers.
The position data, signal data and data gathered from peers is then used by the swarm search algorithm to compute the next waypoint. The waypoint and the current position data is used by the navigation algorithm to compute the robot velocities as shown in Alg.~\ref{alg:swarm-mission}. The robot nodes publish the robot velocities, which are then subscribed by the epuck nodes; converted to motor velocities and sent to the epucks.

ROS executes each node in parallel. Since we initialize a node for each robot (robot-node), decision making and navigation is executed in parallel for all the robots within the central device. The outcome of this is a parallel and asynchronous implementation that is flexible to be used with different multi-robot search algorithms.

\subsection{Sound Signal Observation: Data calibration}
\label{sec:sound-source-cal}
The sound source calibration technique within our software code pipeline aims to convert the relatively noisy transducer data from a quad-set of radially distributed microphones around the robots to a scalar value. For measurements taken at time $t$, this scalar value defines the signal observation -- in terms of sound pressure level (spl) -- associated with the approximate geometric center of the robot in ($x,y$) space at time $t$. The workflow of this code framework aims to de-noise and then transform the transducer voltage readings to spl levels associated with the location of the robot's center.

First, the voltage values are passed to an ADC converter, then the signal is de-noised using a built-in digital bandpass filter sampling. 
From this point onwards, the four transducers have produced cleaner voltage samples which are input in packets of four via the ROS subscriber-publisher network of communication. This is enabled by the ROS package provided by \cite{epuck-driver-cpp}. 
However, due to the specific localized geometry of the transducer distribution, the standardized method of approximation via the Law of Large Numbers (LLN) cannot be applied as the small data point size and their unequal distribution case prevent utilization of the true packet mean to serve as the central approximation. Instead, the radial distribution requires weighting of each sensor to ensure appropriate accounting in the centered point value representing the robot's `true' acoustic input value at any time $t$. To achieve this, the geometric location of each transducer was measured relative to the forward-facing orientation perceived from the top view of the e-puck robot. The four transducers were located at 3.5$^\circ$, 75$^\circ$, 285$^\circ$ and 180$^\circ$ with $\pm 1^\circ$ tolerances.
Thus we can use the following expression to compute the weighted approximation of an acoustic sample positioned over the geometric center of a robot:
\begin{equation}
\phi = \sum_{k=0}^{3}\dfrac{{\text{sens}[k]\times r}/({r-d_k})}{\frac{2r}{r-d_0}+\frac{r}{r-d_2}+\frac{r}{r-d_3}}
\end{equation}

{\footnotesize \noindent Where: $r=3.5\pm 1$: Radial Measurement of Robot Body; \\$d=[0.35,0.35,1.25,0.2]\pm [.1,.1,.1,.1]$: Inward Inset Dimension of sensors based on angular location; \\sens$[k]$ = Packetized Input Value at given location of the robot.}


\subsection{Data Filtering}
\label{sec:data-filter}
The signal data gathered by the robots along the trajectory is noisy mainly because of two reasons:
\textbf{a)} Sensor imperfections; \textbf{b)} Environment noise -- this includes sound noise generated by robot movements and reflections commonplace in such an indoor environment. 
The noisy data must be filtered before it is fed to the search algorithm. This is performed using two levels of filtering. 
{\textit{Outlier removal}:} Removal of outliers is done based on the experimental observations and should be modified as per the sensor-quality and the environment being used by any other user. Arbitrarily high values were observed often, likely attributed to sensing imperfections. Since the signal is expected to follow a trend over space, we take a window of $n$ consecutive observations ($(\phi_k)_{k=1}^{n}$). After obtaining the median, a threshold, $\phi_{\text{T}}=\texttt{median}((\phi_k)_{k=1}^{n}) + n$, is calculated and signal values above the threshold are discarded. A smoothing filter is applied to the observations obtained after outlier removal. 

\section{Swarm Search Algorithms} \label{sec:swarm-search-algos}
In this section, we first provide a brief review of the existing literature in swarm search. Following that, we give the mathematical description of the algorithms that we deployed in our physical setup. Code implementations of these algorithms are available in our github repository \cite{adams_signal_search_testbed}.

\subsection{Swarm Search Algorithms} \label{subsec:swarm search-review}
Swarm Search Algorithms mainly differ in how they perform the informed decision-making of the next waypoint. Broadly, such algorithms can be divided into model-based and model-free methods. 
Model-Based methods employ techniques to create a model of the environment and use a sampling method to sample the next waypoint, and then update the environment model, and repeat this process. Various methods used to create a model of the environment include Hidden Markov Model \cite{farrell2003plume}, partial differential equations aided gas-source localization \cite{wiedemann2019model}, Bayesian-model \cite{ghassemi2020extended}, \cite{ghassemi2022penalized}, and Particle Filters \cite{charrow2014cooperative}. The advantage of model-based methods is that they are explainable, since decision making is governed by a belief model of the signal environment. However, these methods usually face the curse of dimensionality and hence do not scale well. 

Model-free methods, on the other hand, use some form of search heuristics or rules that directly use the signal observations collected, without building an explicit belief model of the signal distribution over space. Nature-Inspired, population based swarm-intelligence (SI) algorithms is the most dominant class of such model-free methods. Generally speaking, they consider each robot in the team or swarm as a particle in a population that moves based on some motion heuristics. 
Wang et al. \cite{wang2022odor} provides a review of swarm intelligence (SI) algorithms used for odor source localization. Dadgar et al. \cite{DADGAR201662} lists a brief description of the formulations of the existing literature on robotic implementations of particle swarm optimization, glowworm swarm optimization, foraging swarm grouping, bidirectional random walk, and bee swarm optimization. Many of these include variants of the concept underlying Particle Swarm Optimization (PSO) \cite{feng2020experimental,li2008probability,yan2018modified,DADGAR201662}. As they obviate the need to construct and refine belief models of the signal environment, these model-free methods are more amenable to scaling with the size of robot swarm. On the flip side they tend to be more black-box and provide limited mathematical guarantees (e.g., optimality) regarding the decisions made. 

\subsection{Algorithms deployed in our physical setup}
To demonstrate the utility of our physical setup and the associated easy-to-use software pipeline, we evaluate three different swarm search algorithms in physical experiments. Summary descriptions of the algorithms are given here. 

\textbf{\textit{1) Bayes-Swarm:}}
Bayes-Swarm is a recent model-based swarm search algorithm that has been shown to exceed existing methods \cite{ghassemi2020extended,ghassemi2022penalized}. At each waypoint-planning instance, Gaussian Process \cite{williams1995gaussian} is used to fit the measured signal data and create the belief model of the signal environment. Subsequently, a custom acquisition function is optimized to determine the new waypoint $x_r^{t+1}$ for robot $r$, as expressed below. 
\begin{equation}\label{eq:bayes-eqn}
    \begin{aligned}
  & x_r^{t+1} = \underset{x} {\arg \max}(\alpha \Omega_r + (1 - \alpha)\beta \Sigma_r)\Gamma_r \\
    & {\text{where}~} 0 \leq l_s^{k_r} = ||x-x_r^{t}|| \leq VT
     \end{aligned}
\end{equation}
The constraint ensures that the distance between consecutive waypoints is $\le$ the product of the robot's nominal velocity $V$ and time step $T$ between waypoint decisions.
Here $\alpha$ and $\beta$  are weighting coefficients, and $\Omega$, $\Sigma$ and $\Gamma$ are respectively the mean source seeking, knowledge uncertainty and penalty terms; further details can be found in \cite{ghassemi2022penalized}.


\textbf{\textit{2) Particle Swarm Optimization (PSO):}}
PSO \cite{kennedy1995particle} is one of the most popular metaheuristics algorithms for search over continuous spaces, which has also been adapted in various forms to perform embodied multi-robot or swarm search. The primary search process in PSO can be expressed in terms of the equations used to update the location of each particle at each iteration, as given by:
\begin{equation} \label{eqn:pso-eqn}
\begin{aligned}
v_i(t) &= \alpha v_i(t-1) + r_1 \beta_l (p_{\text{b}} - x_i(t)) + r_2 \beta_g (g_{\text{b}} - x_i(t)) \\
x_i(t) &= x_i(t-1) + v_i(t)
\end{aligned}
\end{equation}
%
Here, $p_{\text{b}}$ is the position with best function value in the particle's (robot's) own history, and 
$g_{\text{b}}$ is the position with best function value among all the particles (i.e., all robots in the team). $\beta_l$ and $\beta_g$ are scaling factors, heuristically set by domain expert and $r_1, r_2$ are randomly drawn between $0$ and $1$. 
Unlike in non-embodied optimization search, for multi-robot search, signal measurements (function values) sampled across entire trajectory (as opposed to just at waypoints) are used to calculate $p_{\text{best}}$ and $g_{\text{best}}$ in PSO.

\textbf{\textit{3) Random Walk:}}
At each waypoint deciding instance, a robot randomly generates a waypoint within the search arena boundaries, and travels to that point. Information exchange across peers is not used in this case. 

\section{Experiments} \label{sec:num-exps}
\subsection{Physical Experiments}
\begin{figure}[h]
\begin{minipage}[h]{0.95\linewidth}
\centering
\includegraphics[width=0.75\linewidth]{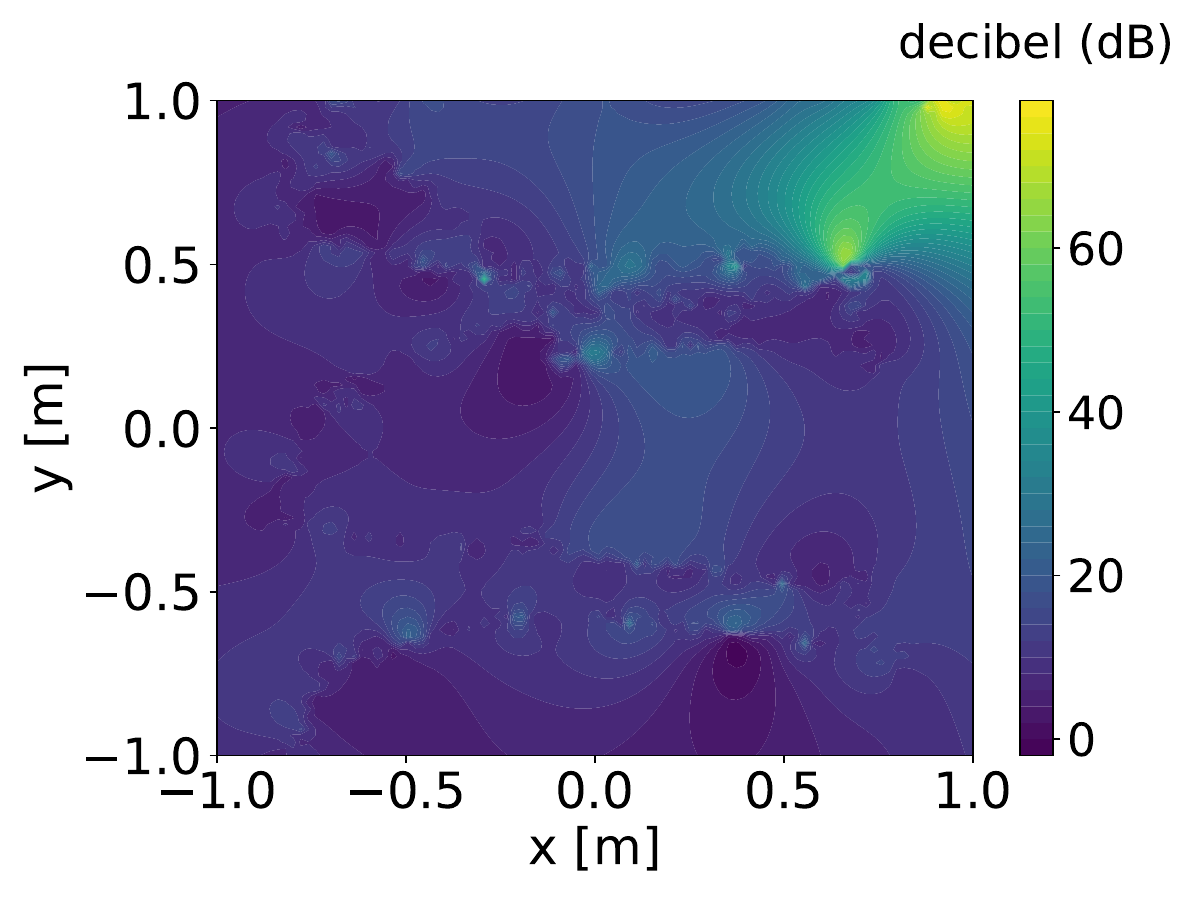}
\subcaption{Real world signal distribution} \label{subfig:real-dist}
\end{minipage}
\begin{minipage}[h]{0.95\linewidth}
\centering
\includegraphics[width=0.75\linewidth]{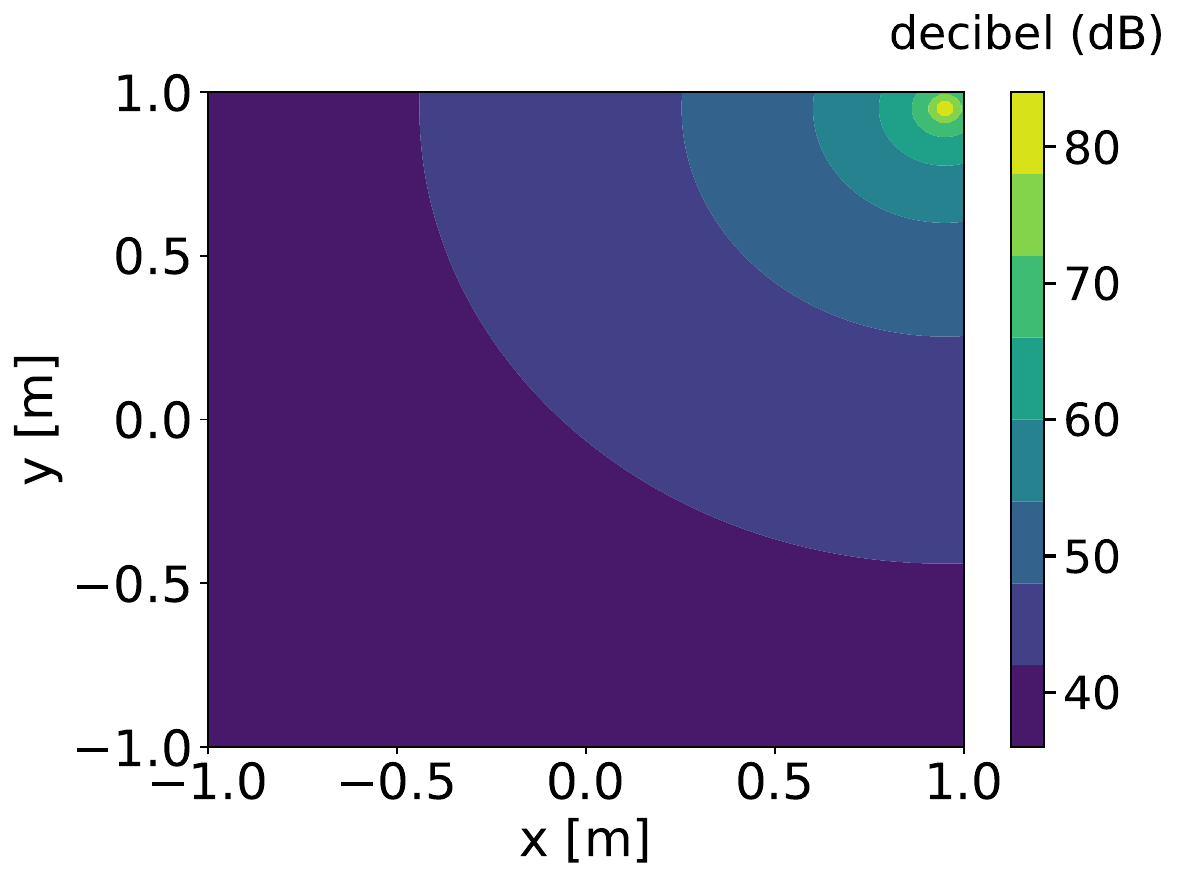}
\subcaption{Signal distribution in simulation} \label{subfig:sim-dist}
\end{minipage}
    \caption{Example signal distribution in the physical experiment and in simulation, realized by an acoustic source at (0.95 m, 0.95 m), used for initial acoustic environment assessment}
    \label{fig:sim-real-gap}
\end{figure}
All experiments are conducted using 4 robots. 
In the physical setup, to assess the nature of acoustic environment as experienced by the robots, we first place our sound signal source (Sec.~\ref{sec:hardware}) at $(0.95, 0.95)$m. We then scan the environment with an e-puck using a pre-coded trajectory. After filtering the data, the resulting contour plot of the signal is shown in Fig.~\ref{subfig:real-dist}. RBF interpolator from scipy \cite{2020SciPy-NMeth} with a linear kernel was used to derive the contour map. 

Our design of experiments considers variation in the source location and starting position of the robots relative to the source. Note that while the source is exactly the same, these source locations in the experiment cases are different from that shown in Fig.~\ref{subfig:real-dist}.
These experiment cases are listed in Table \ref{tab:bayes-swarm-pso-rw}. Note that since the focus of this paper is to present this new physical setup and software pipeline, the experiments with the different search algorithm are mainly demonstrative; they are not intended to serve as an in-depth compare/contrast evaluation of those algorithms (not within the scope of this paper). Hence, not all the three swarm algorithms are subjected to all the experimental cases. Bayes-Swarm is the one that provides the most promising performance in physical experiments, and is thus analyzed over more number of cases than the others. 

The mission is terminated if any one of the robots finds the source location (within a tolerance of 0.2 m) or the mission time exceeds $300$ seconds. The latter termination criterion indicates a failed search mission. To account for the stochasticity in robot-movements, environment noise, the mission time is found to vary when the same experiment case is repeated with the same algorithm. Hence each case is run 5 times, and the statistical results are reported in Table \ref{tab:bayes-swarm-pso-rw}. We have provided the minimum and maximum mission-completion time across the runs.

\begin{table}[h]
    \centering
    \begin{tabular}{|c|c|c|c|c|c|}
        \hline
        Alg & Start(m) & Source(m) & min(s) & max(s) \\
        \hline
        BS  & (0.0, 0.0) & (0.85, 0.85) & 14 & 251 \\
        \hline
        BS  & (0.85, -0.85) & (0.85, 0.85) & 11.39 & 270.13 \\ \hline
        BS  & (0.85, -0.85) & (-0.85, 0.85) & 300 & 300 \\
        \hline
        RW  & Random & (0.85, 0.85) & 30 & 300 \\ \hline
        PSO & (0.0, 0.0) & (0.85, 0.85) & 300 & 300 \\
        \hline
    \end{tabular}
    \caption{Evaluation of Bayes-Swarm (BS), Particle Swarm Optimization (PSO), Random Walk (RW). Start refers to robot start position and source refers the location of the signal source (sec.~\ref{sec:hardware}). Min(s) and max(s) refers to the minimum and maximum mission completion time observed across 5 runs for each case}
    \label{tab:bayes-swarm-pso-rw}
\end{table}

From Table \ref{tab:bayes-swarm-pso-rw}, we observe that  for the first two cases, Bayes-Swarm \cite{ghassemi2022penalized} was able to find the signal source before the mission time was over in all 5 runs, and recorded promising minimum search times. In these two cases Bayes-Swarm was able to escape local optima (initial robot positions being close to the source location) and deal with the noisiness to some extent. For the third case, where the robot starting position and the source location are at diagonally opposite ends of the search arena (($0.85, -0.85$) vs. ($-0.85, 0.85$)), Bayes-Swarm failed to find the source across 5 runs. For the single cases on which PSO and Random Walk algorithms were tested, PSO failed to find the source across all 5 runs. We observed that the robots got stuck at a local optimum near the start point. Random Walk found the source in 3/5 runs (Table \ref{tab:bayes-swarm-pso-rw}). However, performance of random walk is chance driven, and serves as a lower baseline in comparison.

\subsection{Simulation experiments \& Sim-to-Real Discussion}
We used the simulation environment provided by \cite{ghassemi2022penalized} to replicate the physical setup. In the simulation environment, the robots are modeled as dots and the signal is modeled using an analytical function. 
The acoustic signal source is modeled as a point source and the spl measurements are calculated using the following equation \cite{kinsler2000fundamentals}:

\begin{equation}
\begin{aligned}
f &= 20 \cdot \log_{10} \left( \frac{p}{p_{\text{ref}}} \right), \\
p &= \frac{p_0}{\sqrt{2} \cdot r}, \quad 
r = \max\left( \lVert \mathbf{x} - \mathbf{x}_s \rVert,\, 0.01 \right)
\end{aligned}
\end{equation}

$f$ gives the spl measurement at a location $\mathbf{x}$, for the acoustic point source placed at $\mathbf{x}_s$. $p_{\text{ref}}$ is the referenced effective pressure amplitude, equal to $20 \mu$ pa \cite{kinsler2000fundamentals}, $p$ is the measured pressure amplitude of the sound wave, $r$ is the norm of the distance between the source ($\mathbf{x}_s$) and the location ($\mathbf{x}$). We used phone as the signal source in our physical experiments (sec.~\ref{sec:hardware}). We set $p_0$ to $0.00495$ Pa, pressure corresponding to $85$dB, the maximum spl measurement observed in the physical experiments.

\begin{table}[h]
    \centering
    \begin{tabular}{|c|c|c|c|c|c|}
        \hline
        Alg & Start(m) & Source(m) & min(s) & max(s)\\
        \hline
        BS  & (0.0, 0.0) & (0.85, 0.85) & 27.39 & 27.39 \\
        \hline
        BS  & (0.85, -0.85) & (0.85, 0.85) & 34.31 & 34.31 \\ \hline
        BS  & (0.85, -0.85) & (-0.85, 0.85) & 66 & 127.96 \\
        \hline
        RW  & Random & (0.85, 0.85) & 32 & 274 \\ \hline
        PSO & (0.0, 0.0) & (0.85, 0.85) & 300 & 300 \\
        \hline
    \end{tabular}
    \caption{Simulation results of Bayes-Swarm (BS), Particle Swarm Optimization (PSO), Random Walk (RW). Start refers to robot start position and source refers the location of the signal source (sec.~\ref{sec:hardware}). Min(s) and max(s) refers to the minimum and maximum mission completion time observed across 5 runs for each case}
    \label{tab:sim-results}
\end{table}

\begin{table}[h]
    \centering
    \begin{tabular}{|c|c|c|c|c|c|}
        \hline
         Alg & Start(m) & Source(m) & min(s) & max(s) \\
        \hline
        BS  & (0.0, 0.0) & (0.85, 0.85) & 30 & 49 \\
        \hline
        BS  & (0.85, -0.85) & (0.85, 0.85) & 34.31 & 38.31 \\ \hline
        BS  & (0.85, -0.85) & (-0.85, 0.85) & 126 & 300 \\
        \hline
        RW  & Random & (0.85, 0.85) & 34 & 300 \\ \hline
        PSO & (0.0, 0.0) & (0.85, 0.85) & 300 & 300 \\
        \hline

    \end{tabular}
    \caption{Simulation results with noisy measurements. Start refers to robot start position and source refers the location of the signal source (sec.~\ref{sec:hardware}). Min(s) and max(s) refers to the minimum and maximum mission completion time observed across 5 runs for each case}
    \label{tab:sim-results-noisy}
\end{table}

Table \ref{tab:sim-results} shows the simulation results for all the five test-cases conducted on the physical setup. 
The robots could not complete the mission with PSO even in simulation environment. This is because of the nature of the signal. As can be seen in Fig.~\ref{subfig:sim-dist}, the signal values do not change significantly over small change in locations. Therefore, the following terms in eqn.~\ref{eqn:pso-eqn}, current location ($x_i(t)$), local best ($p_b$) and global best ($g_b$) were often the same leading the robots to get stuck after their first heuristic decision. In Bayes-Swarm, a model of the environment is created and the next waypoint is planned considering exploitation and exploration. Hence the robots were able to localize the source with Bayes-Swarm.

The mission time of Bayes-Swarm has less variance than the physical experiments. The comparison of minimum mission time for the first two cases in physical setup (Table \ref{tab:bayes-swarm-pso-rw}) versus in simulation (Table \ref{tab:sim-results}) reveals that the minimum mission completion time is greater than that observed in the physical experiments.
This is because of the robot speed. While the robot speed in physical experiments was the same as simulation, the robot speeds often change during the mission because of the uneven terrain. The maximum mission time is however less than that observed in real world. This is because of two factors 1) The real world signal is too noisy as compared to the simulation (shown in Fig.~\ref{subfig:sim-dist}, \ref{subfig:real-dist}). Therefore, the robots often got stuck in local optima, and since multiple robots got stuck there, they need to engage in negotiation procedures which cause a delay in the mission. 2) Noisy measurements: Inspite of filtering techniques used, the noisy sensor measurements cause the model based methods to give inaccurate models which affects the mission time. We used the first factor to improve Bayes-Swarm performance which will be further discussed in next section. Next, we tried to address the second factor by conducting a new set of simulation experiments with Bayes-Swarm by including noisy measurements in the simulation. Table \ref{tab:sim-results-noisy} shows the results of Bayes-Swarm with noisy measurements. The minimum and maximum mission time of Bayes-Swarm saw a slight increase for the first two cases, whereas for the third case the robots were not able to find the source in the time limit (as observed in physical experiments results, Table \ref{tab:bayes-swarm-pso-rw}).

\subsection{Using the Setup to Improve Algorithm Performance}
Factors such as systemic uncertainty in robots (battery state, motors etc.), uncertainty in robots' movement, communication latency, uncertainty in sensor measurements, environment interferences, all lead to search outcomes (and usually loss in performance) in the physical experiments that deviate from those achieved in simulation.  


Now, in the first set of experiments with Bayes-swarm (shown in Table \ref{tab:bayes-swarm-pso-rw}), we observed that robots in the team kept converging to similar locations, which also triggered (time-wasting) collision evasion procedures among them (often not accounted for in the simulated deployment). This behavior is supposed to be mitigated by the penalty term that hinges on the Lipschitz continuity condition of the signal function to penalize interactions between different robots planned waypoints. This condition is however not met in noisy physical environments, making the penalty term ineffective. Hence, to address the performance loss due to this sim-to-real gap, we explored the following modified version of the penalty term: $\gamma = 1 + {||x - x_p^{-r}||}/\left(
    {1 + ||\mathbf{x_\text{U}} - \mathbf{x_\text{L}}||}\right)$; here $x_p^{-r}$ is the peers' planned waypoint. 

In addition, to promote increased exploration and get any robot $r$ out of the many potential local optima (encountered due to noisy signal distribution experienced in practice, but not in simulation), we employ this modified condition in BS: $ \texttt{if} \hspace{2mm} ||x_p^{r} - x_c^{r}|| < \epsilon$,\hspace{1mm} $ \alpha 
 \gets 0.1$. Here, $x_p^{r}$ is the next planned waypoint, $x_c^{r}$ is current-position, $\epsilon \gets 0.1$. The resulting Bayes-Swarm implementation with modified penalty and exploration condition is called BS-P. 

Table \ref{tab:new-old-bs} shows the performance of the modified Bayes-Swarm (BS-P) implementation on the same cases in which the original BS was executedrun. We see now that robots are able to localize the signal source more consistently, (more robust across runs), reporting low mission completion times across the 5 runs in each of the first two cases. Moreover, now the BS-P is able to succeed in finding the source in the challenging third case where the original BS had failed. Such adaptations become possible only with the availability of an experimental setup and software pipeline as presented here.

\begin{table}[h]
    \centering
    \begin{tabular}{|c|c|c|c|c|c|}
        \hline
         Start(m) & Source(m) & min(s) & max(s) \\
        \hline
        (0.0, 0.0) & (0.85, 0.85) & 17.29 & 18.696 \\
         \hline
         (0.0, 0.0) & (-0.85, 0.85) & 14 & 17.69 \\
        \hline
        (0.85, -0.85) & (-0.85, 0.85) & 54.75 & 190.25 \\
        \hline
    \end{tabular}
    \caption{Mission completion time for BS-P: Bayes-Swarm with new penalty factor term. Start refers to robot start position and source refers the location of the signal source (sec.~\ref{sec:hardware}). Min(s) and max(s) refers to the minimum and maximum mission completion time observed across 5 runs for each case}
    \label{tab:new-old-bs}
\end{table}

\begin{figure}[h]
\begin{minipage}{0.49\linewidth}
    \includegraphics[width=\linewidth]{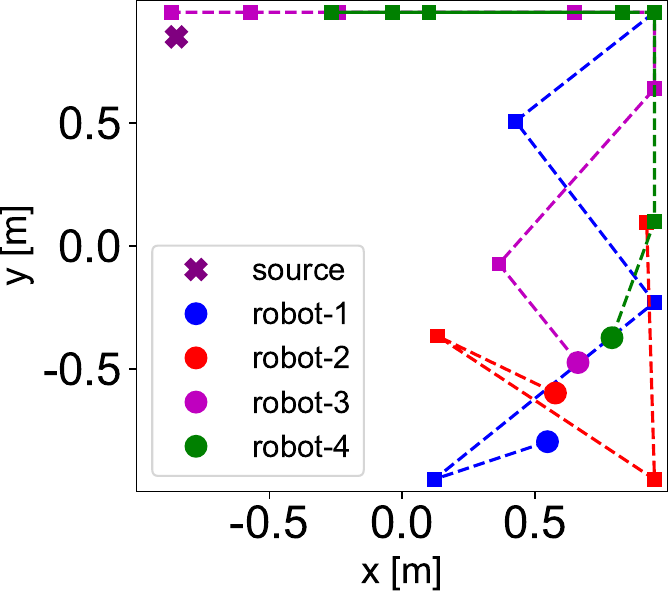}
\subcaption[a]{Robot trajectories}
\end{minipage}%
\begin{minipage}{0.49\linewidth}
    \includegraphics[width=\linewidth]{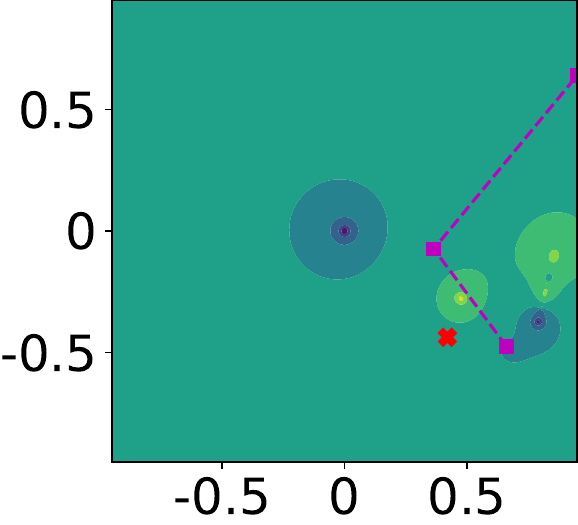}
\subcaption[b]{Robot-3 at 9s}
\end{minipage}%

\begin{minipage}{0.497\linewidth}
    \includegraphics[width=\linewidth]{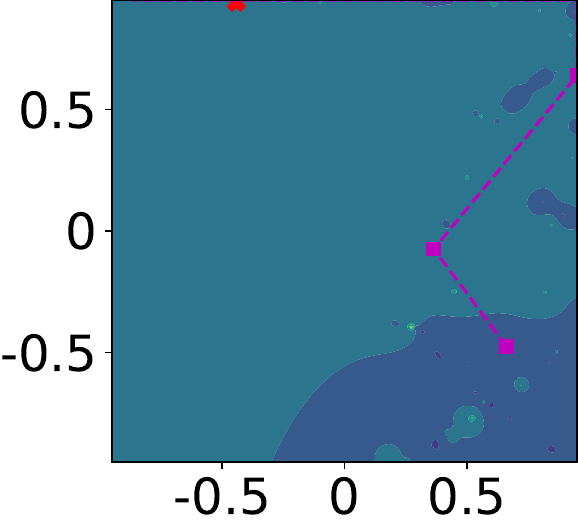}
\subcaption[d]{Robot-3 at 57s, source found}
\end{minipage}%
\begin{minipage}{0.52\linewidth}\includegraphics[width=\linewidth]{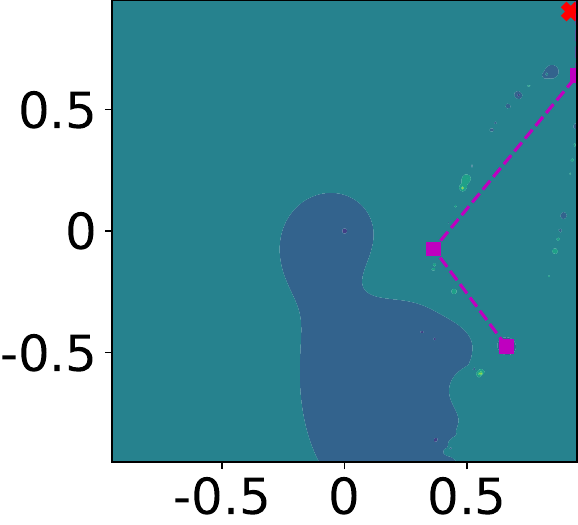}
\subcaption[c]{Robot-3 at 36s}
\end{minipage}%
\caption{Example BS-P mission,: (a) Robot trajectories and true source; b) $\to$ d) $\to$ c) Evolution of belief model for Robot-3, which finds the source at t=$57$ sec. Red cross shows the expected source location based on then current GP model.} \label{fig:gp-contours}
\end{figure}

Figure~\ref{fig:gp-contours} shows an example of the over-time evolution of trajectories of four robots, and that of the GP belief model (mean prediction) for one of the robots, using the new BS-P algorithm in this case.

\section{Conclusion} \label{sec:concl}
In this paper, we presented a new experimental setup to evaluate optimization based multi-robot search algorithms for signal source localization implemented on physical robots. The setup uses a readily available and safe acoustic source, and integrates an open-source software stack for physical robots (e-pucks in our case), a motion-capture system, and readily integrates with search/navigation algorithm using a ROS framework. 
To demonstrate this experimental setup and associated software pipeline, we used it to evaluate model free (PSO), model based (Bayes-Swarm) and Random Walk search algorithms. Experiments are conducted with four robots. Uncertainties and resulting poor signal-to-noise ratio in the physical environment were observed to have a significant impact on search performance. However, the Bayes-Swarm algorithm still provided 100\% success rate in finding the acoustic source in 2 out of 3 experiments, while also recording a relatively fast mission completion in the successful cases. In contrast, PSO was observed to fail in finding the source in each case. Based on the experimental observations, penalization formulations in the Bayes-Swarm algorithm were also refined, resulting in further improvement in its search performance. In its current form, the hardware/software pipeline does not account for obstacle avoidance, and is mostly tied to the specific robotic platform used. Alleviation of these constraints, and development of more systematic processes for algorithm benchmarking and reduction of sim-to-real gaps, in future work would further enhance the potential adoption of this experimental setup and protocol across the robotics community. 

\bibliographystyle{ieeetr}
\bibliography{references}

\begin{thebibliography}{10}

\bibitem{tadokoro2019disaster}
S.~Tadokoro, {\em Disaster robotics: results from the ImPACT tough robotics challenge}, vol.~128.
\newblock Springer, 2019.

\bibitem{baetz2009mobile}
W.~Baetz, A.~Kroll, and G.~Bonow, ``Mobile robots with active ir-optical sensing for remote gas detection and source localization,'' in {\em 2009 IEEE international conference on robotics and automation}, pp.~2773--2778, IEEE, 2009.

\bibitem{charrow2014cooperative}
B.~Charrow, N.~Michael, and V.~Kumar, ``Cooperative multi-robot estimation and control for radio source localization,'' {\em The International Journal of Robotics Research}, vol.~33, no.~4, pp.~569--580, 2014.

\bibitem{viseras2016decentralized}
A.~Viseras, T.~Wiedemann, C.~Manss, L.~Magel, J.~Mueller, D.~Shutin, and L.~Merino, ``Decentralized multi-agent exploration with online-learning of gaussian processes,'' in {\em 2016 IEEE international conference on robotics and automation (ICRA)}, pp.~4222--4229, IEEE, 2016.

\bibitem{behjat2021learning}
A.~Behjat, H.~Manjunatha, P.~K. Kumar, A.~Jani, L.~Collins, P.~Ghassemi, J.~Distefano, D.~Doermann, K.~Dantu, E.~Esfahani, {\em et~al.}, ``Learning robot swarm tactics over complex adversarial environments,'' in {\em 2021 international symposium on multi-robot and multi-agent systems (MRS)}, pp.~83--91, IEEE, 2021.

\bibitem{8490881}
S.~V. Sibanyoni, D.~T. Ramotsoela, B.~J. Silva, and G.~P. Hancke, ``A 2-d acoustic source localization system for drones in search and rescue missions,'' {\em IEEE Sensors Journal}, vol.~19, no.~1, pp.~332--341, 2019.

\bibitem{lochmatter2009understanding}
T.~Lochmatter and A.~Martinoli, ``Understanding the potential impact of multiple robots in odor source localization,'' in {\em Distributed Autonomous Robotic Systems 8}, pp.~239--250, Springer, 2009.

\bibitem{tan2013research}
Y.~Tan and Z.-y. Zheng, ``Research advance in swarm robotics,'' {\em Defence Technology}, vol.~9, no.~1, pp.~18--39, 2013.

\bibitem{wang2022odor}
J.~Wang, Y.~Lin, R.~Liu, and J.~Fu, ``Odor source localization of multi-robots with swarm intelligence algorithms: A review,'' {\em Frontiers in Neurorobotics}, vol.~16, p.~949888, 2022.

\bibitem{DADGAR201662}
M.~Dadgar, S.~Jafari, and A.~Hamzeh, ``A pso-based multi-robot cooperation method for target searching in unknown environments,'' {\em Neurocomputing}, vol.~177, pp.~62--74, 2016.

\bibitem{sinha2018convergence}
A.~Sinha and R.~K. Mishra, ``Convergence of multi-agent systems to unknown source of an odor,'' in {\em 2018 3rd international conference for convergence in technology (I2CT)}, pp.~1--6, IEEE, 2018.

\bibitem{jing2021recent}
T.~Jing, Q.-H. Meng, and H.~Ishida, ``Recent progress and trend of robot odor source localization,'' {\em IEEJ Transactions on Electrical and Electronic Engineering}, vol.~16, no.~7, pp.~938--953, 2021.

\bibitem{ghassemi2020extended}
P.~Ghassemi and S.~Chowdhury, ``An extended bayesian optimization approach to decentralized swarm robotic search,'' {\em ASME Journal of Computing and Information Science in Engineering}, vol.~20, no.~5, p.~051003, 2020.

\bibitem{ghassemi2022penalized}
P.~Ghassemi, M.~Balazon, and S.~Chowdhury, ``A penalized batch-bayesian approach to informative path planning for decentralized swarm robotic search,'' {\em Autonomous Robots}, vol.~46, no.~6, pp.~725--747, 2022.

\bibitem{mokhtarian2024survey}
A.~Mokhtarian, J.~Xu, P.~Scheffe, M.~Kloock, S.~Sch{\"a}fer, H.~Bang, V.-A. Le, S.~Ulhas, J.~Betz, S.~Wilson, {\em et~al.}, ``A survey on small-scale testbeds for connected and automated vehicles and robot swarms: A guide for creating a new testbed,'' {\em IEEE Robotics \& Automation Magazine}, 2024.

\bibitem{ouguz2024open}
S.~O{\u{g}}uz, M.~K. Heinrich, M.~Allwright, W.~Zhu, M.~Wahby, E.~Garone, and M.~Dorigo, ``An open-source uav platform for swarm robotics research: Using cooperative sensor fusion for inter-robot tracking,'' {\em IEEE access}, vol.~12, pp.~43378--43395, 2024.

\bibitem{wiedemann2019model}
T.~Wiedemann, D.~Shutin, and A.~J. Lilienthal, ``Model-based gas source localization strategy for a cooperative multi-robot system—a probabilistic approach and experimental validation incorporating physical knowledge and model uncertainties,'' {\em Robotics and Autonomous Systems}, vol.~118, pp.~66--79, 2019.

\bibitem{quigley2009ros}
M.~Quigley, K.~Conley, B.~Gerkey, J.~Faust, T.~Foote, J.~Leibs, R.~Wheeler, A.~Y. Ng, {\em et~al.}, ``Ros: an open-source robot operating system,'' in {\em ICRA workshop on open source software}, vol.~3, p.~5, Kobe, Japan, 2009.

\bibitem{adams_signal_search_testbed}
``adamslab-ub/{M}ulti\_{R}obot\_{S}ignal\_{S}earch\_{T}estbed.'' \url{https://github.com/adamslab-ub/Multi_Robot_Signal_Search_Testbed.git}.

\bibitem{alkilabi2017cooperative}
M.~H.~M. Alkilabi, A.~Narayan, and E.~Tuci, ``Cooperative object transport with a swarm of e-puck robots: robustness and scalability of evolved collective strategies,'' {\em Swarm intelligence}, vol.~11, pp.~185--209, 2017.

\bibitem{harmanda2021development}
T.~T. Harmanda, M.~K. Hardhienata, and K.~Priandana, ``Development of multi-robot systems using particle swarm optimization algorithm for task allocation,'' in {\em 2021 IEEE Region 10 Symposium (TENSYMP)}, pp.~1--8, IEEE, 2021.

\bibitem{florea2020overview}
A.~G. Florea and C.~Buiu, ``An overview of swarm robotics,'' {\em Robotic Systems: Concepts, Methodologies, Tools, and Applications}, pp.~58--69, 2020.

\bibitem{viconMotionCapture}
``{M}otion {C}apture for the {E}ngineering {I}ndustry | {V}icon --- vicon.com.'' \url{https://www.vicon.com/applications/engineering/}.
\newblock [Accessed 16-09-2024].

\bibitem{epuck-driver-cpp}
``{G}it{H}ub - gctronic/epuck\_driver\_cpp: {E}-puck {R}{O}{S} node based on roscpp.'' \url{https://github.com/gctronic/epuck_driver_cpp}.

\bibitem{vicon}
``{G}it{H}ub - ethz-asl/vicon\_bridge.'' \url{https://github.com/ethz-asl/vicon_bridge}.

\bibitem{farrell2003plume}
J.~A. Farrell, S.~Pang, and W.~Li, ``Plume mapping via hidden markov methods,'' {\em IEEE Transactions on Systems, Man, and Cybernetics, Part B (Cybernetics)}, vol.~33, no.~6, pp.~850--863, 2003.

\bibitem{feng2020experimental}
Q.~Feng, H.~Cai, Y.~Yang, J.~Xu, M.~Jiang, F.~Li, X.~Li, and C.~Yan, ``An experimental and numerical study on a multi-robot source localization method independent of airflow information in dynamic indoor environments,'' {\em Sustainable Cities and Society}, vol.~53, p.~101897, 2020.

\bibitem{li2008probability}
F.~Li, Q.-H. Meng, S.~Bai, J.-G. Li, and D.~Popescu, ``Probability-pso algorithm for multi-robot based odor source localization in ventilated indoor environments,'' in {\em Intelligent Robotics and Applications: First International Conference, ICIRA 2008, Wuhan, China, October 15-17, 2008, Proceedings, Part I 1}, pp.~1206--1215, Springer, 2008.

\bibitem{yan2018modified}
Y.~Yan, R.~Zhang, J.~Wang, and J.~Li, ``Modified pso algorithms with “request and reset” for leak source localization using multiple robots,'' {\em Neurocomputing}, vol.~292, pp.~82--90, 2018.

\bibitem{williams1995gaussian}
C.~Williams and C.~Rasmussen, ``Gaussian processes for regression,'' {\em Advances in neural information processing systems}, vol.~8, 1995.

\bibitem{kennedy1995particle}
J.~Kennedy and R.~Eberhart, ``Particle swarm optimization,'' in {\em Proceedings of ICNN'95-international conference on neural networks}, vol.~4, pp.~1942--1948, ieee, 1995.

\bibitem{2020SciPy-NMeth}
P.~Virtanen, R.~Gommers, T.~E. Oliphant, M.~Haberland, T.~Reddy, D.~Cournapeau, E.~Burovski, P.~Peterson, W.~Weckesser, J.~Bright, S.~J. {van der Walt}, M.~Brett, J.~Wilson, K.~J. Millman, N.~Mayorov, A.~R.~J. Nelson, E.~Jones, R.~Kern, E.~Larson, C.~J. Carey, {\.I}.~Polat, Y.~Feng, E.~W. Moore, J.~{VanderPlas}, D.~Laxalde, J.~Perktold, R.~Cimrman, I.~Henriksen, E.~A. Quintero, C.~R. Harris, A.~M. Archibald, A.~H. Ribeiro, F.~Pedregosa, P.~{van Mulbregt}, and {SciPy 1.0 Contributors}, ``{{SciPy} 1.0: Fundamental Algorithms for Scientific Computing in Python},'' {\em Nature Methods}, vol.~17, pp.~261--272, 2020.

\bibitem{kinsler2000fundamentals}
L.~E. Kinsler, A.~R. Frey, A.~B. Coppens, and J.~V. Sanders, {\em Fundamentals of acoustics}.
\newblock John wiley \& sons, 2000.

\end{thebibliography}
\end{document}